\documentclass[10pt,twocolumn,letterpaper]{article}

\usepackage{cvpr}              
\usepackage{xcolor}
\usepackage{soul}
\usepackage{array}
\newcolumntype{L}[1]{>{\raggedright\arraybackslash}p{#1}}
\newcolumntype{C}[1]{>{\centering\arraybackslash}p{#1}}

\definecolor{cvprblue}{rgb}{0.21,0.49,0.74}
\usepackage[pagebackref,breaklinks,colorlinks,allcolors=cvprblue]{hyperref}

\title{CPSL: Representing Volumetric Video via Content-Promoted Scene Layers}

\author{
Kaiyuan Hu$^{1}$,\;
Yili Jin$^{1}$,\;
Junhua Liu$^{2}$,\;
Xize Duan$^{4}$,\;
Hong Kang$^{1}$,\;
Xue Liu*$^{1,3}$
\\[4pt]
$^1$McGill University\\
$^2$University of Southern California\\
$^3$Mohamed bin Zayed University of Artificial Intelligence\\
$^4$The Chinese University of Hong Kong, Shenzhen
\\[4pt]
\texttt{(kaiyuan.hu, yili.jin, hong.kang)@mail.mcgill.ca} \\
\texttt{xue.liu@mcgill.ca}\\
\texttt{junhual@usc.edu} \\
\texttt{xizeduan@link.cuhk.edu.cn}
}

\begin{document}
\maketitle

\begin{abstract}
Volumetric video enables immersive and interactive visual experiences by supporting free viewpoint exploration and realistic motion parallax. However, existing volumetric representations from explicit point clouds to implicit neural fields, remain costly in capture, computation, and rendering, which limits their scalability for on-demand video and reduces their feasibility for real-time communication. 

To bridge this gap, we propose Content-Promoted Scene Layers (CPSL), a compact 2.5D video representation that brings the perceptual benefits of volumetric video to conventional 2D content. Guided by per-frame depth and content saliency, CPSL decomposes each frame into a small set of geometry-consistent layers equipped with soft alpha bands and an edge–depth cache that jointly preserve occlusion ordering and boundary continuity. These lightweight, 2D-encodable assets enable parallax-corrected novel-view synthesis via depth-weighted warping and front-to-back alpha compositing, bypassing expensive 3D reconstruction. Temporally, CPSL maintains inter-frame coherence using motion-guided propagation and per-layer encoding, supporting real-time playback with standard video codecs. Across multiple benchmarks, CPSL achieves superior perceptual quality and boundary fidelity compared with layer-based and neural-field baselines while reducing storage and rendering cost by several folds. Our approach offer a practical path from 2D video to scalable 2.5D immersive media.

\end{abstract}

\section{Introduction}
Volumetric video (VV) has emerged as a compelling format for immersive media, enabling users to explore dynamic 3D environments with six degrees of freedom (6DoF). Unlike conventional 2D video, which fixes the viewpoint at capture time, volumetric formats preserve spatial structure and enable users to move, rotate, and zoom as if physically present. This unique capability underpins key applications in immersive telepresence, interactive entertainment \cite{jin2023capture}, and telemedicine \cite{wu2024ai}, positioning volumetric video as a cornerstone for next-generation interactive media.

\begin{figure}[t]
  \centering
  \includegraphics[width=\linewidth]{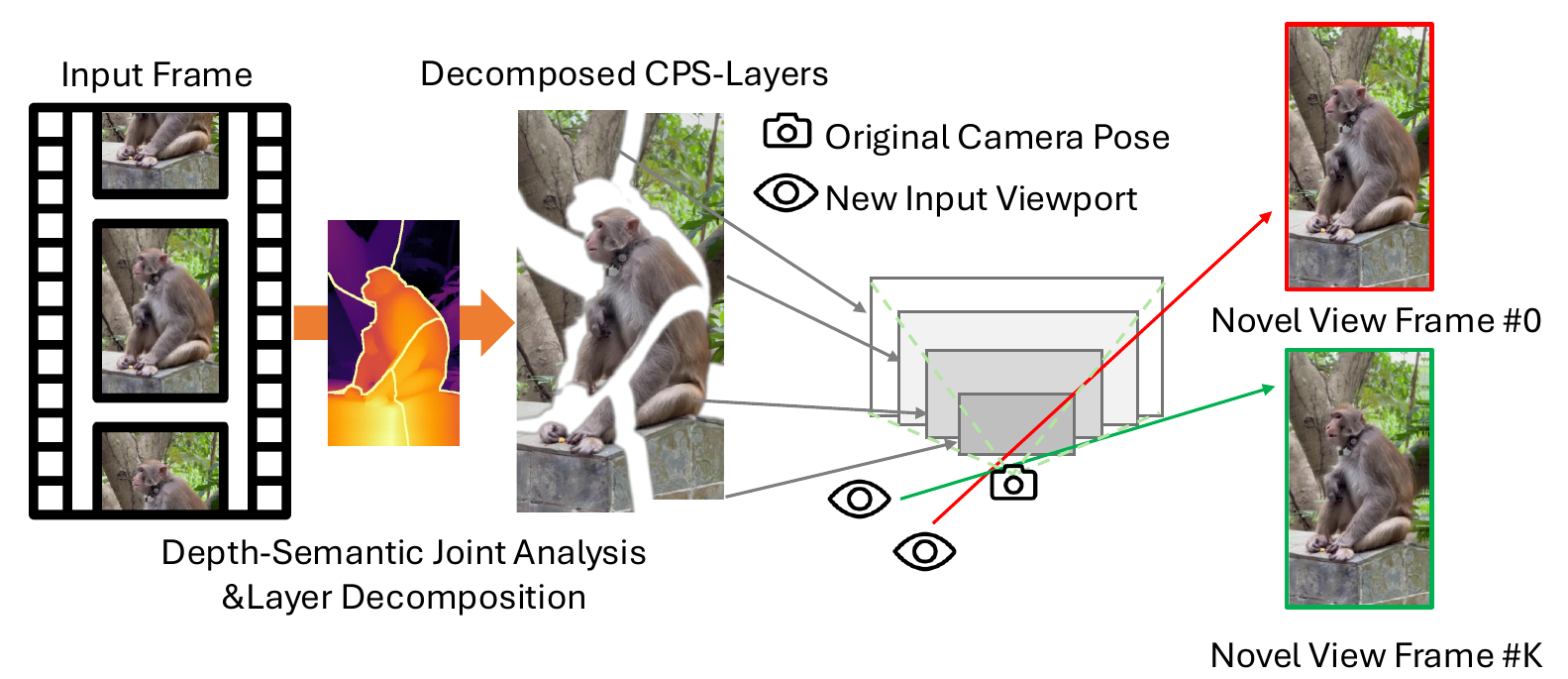}

  \vspace{-2mm} 
  \caption{
  \textbf{Overview of CPSL.}
  From a single 2D video, CPSL constructs a compact set of 
  content-promoted scene layers with boundary-aware geometry.}
  \label{fig:cpsl_overview}
  \vspace{-3mm}
\end{figure}

State-of-the-art volumetric pipelines can be broadly categorized into \textit{explicit}, \textit{implicit}, and \textit{layered} representations. 
\textit{Explicit methods}, such as textured meshes and point clouds, directly encode geometry and texture but require dense capture and incur large storage and bandwidth~\cite{shi2023vpcc,choi2023tmo,hu2025livevv}. 
\textit{Implicit neural fields}, exemplified by NeRF and its dynamic variants~\cite{mildenhall2020nerf,pumarola2021dnerf}, achieve high visual fidelity but suffer from expensive training and slow inference, limiting real-time deployment. 
\textit{Hybrid neural point-based} techniques like 3D Gaussian Splatting (3D-GS)~\cite{kerbl2023gaussiansplatting} strike a balance between explicit geometry and implicit rendering but rely on multi-view supervision and heavy pre-processing. 
More recent variants such as 4D Gaussian Splatting (4D-GS)~\cite{wu20244d,wang2025shape} further enable dynamic 3D scene representation by augmenting Gaussians with temporal components. Beyond these 3D-centric paradigms, \textit{layered 2.5D representations} such as MPI and LDI~\cite{zhang2023structural,zhou2018stereo,tucker2020single,shade1998layered} offer lightweight rendering and compression by encoding depth-ordered semi-transparent planes, yet their fidelity quickly degrades under large viewpoint shifts or dynamic motion, especially in monocular videos. These limitations motivate the search for a more content-adaptive layered form that maintains spatial and temporal coherence across viewpoints and frames.


Despite these advances, immersive video systems remain fundamentally constrained by an inherent balance among \textit{viewing freedom}, \textit{latency}, and \textit{bandwidth cost}. Increasing viewpoint freedom enlarges data volume, while achieving low latency requires substantial computation and limited bitrates reduce geometric and perceptual fidelity. Existing representations navigate this trade-off differently, neural fields favor fidelity but incur heavy optimization, explicit point or mesh based methods render efficiently but rely on dense multi-view capture, and layered 2.5D formats compress well yet degrade under large viewport offsets, therefore none provide a balanced real-time solution. Beyond the limitations of individual representations, system level scalability also poses a constraint: server-side rendering pipelines must process each user’s 6DoF request independently, causing computational load to increase rapidly in multi-user scenarios. This motivates a lightweight, content-adaptive representation that preserves spatial perception while remaining compatible with standard video streaming infrastructure, enabling practical `render-once, stream-many' immersive media with substantially reduced compute and bandwidth requirements.

We propose \textbf{Content-Promoted Scene Layers (CPSL)}, a compact layered representation that explicitly addresses the efficiency–fidelity trade-off in volumetric video. CPSL represents each viewpoint as a small stack of depth-ordered RGBA layers $\{L_k^{(v)}\}$, combining geometric cues from depth and semantic signals for content-aware partitioning. A boundary-aware matting and temporal refinement module preserves edge continuity and cross-frame stability, while each layer remains 2D-encodable for efficient streaming. To allocate resources adaptively, a two-level saliency mechanism ranks viewports and layers by perceptual importance, guiding bitrate and rendering budgets. On the receiver side, a lightweight synthesizer reprojects and composites decoded layers via depth-aware warping and alpha blending. By treating layered 2.5D assets as perceptual and coding units, CPSL achieves compact storage, real-time rendering, and high perceptual fidelity compared to heavier 3D representations.


In summary, this work has the following contributions:
\begin{itemize}
    \item \textbf{Content-Promoted Scene Layers (CPSL).} 
    Compact 2.5D representation that decomposes monocular video into a small set of depth-ordered RGBA layers while preserving parallax, occlusion, and boundary continuity.
    
    \item \textbf{Dynamic Pixel Strip (DPS).}
    View-adaptive screen-space transition that removes cracks and double edges in layered rendering under wide-baseline viewpoint changes.
    
    \item \textbf{Scalability to full-scene volumetric video.}
    Experiments show CPSL naturally adapts to VV and enables real-time playback at very low bitrate with standard 2D codecs—without a specialized volumetric pipeline.
\end{itemize}

\section{Related Work}

\subsection{Representation methods.}
Scene representation lies at the core of view-synthesis research.  
Explicit volumetric formats such as textured meshes and point clouds directly encode geometry and texture but require dense capture and heavy storage~\cite{maglo20153d}, limiting their scalability.  
Layered 2.5D representations, including Multiplane Images (MPI)~\cite{tucker2020single,luvizon2021adaptive,zhang2023structural} and Layered Depth Images (LDI)~\cite{shade1998layered}, model scenes with depth-ordered RGBA planes that can be efficiently rendered and compressed.  
However, conventional layering suffers from cracks, double edges, and halo artifacts when reprojected to wide-baseline viewpoints or dynamic scenes~\cite{muddala2016spatio}.  
Recent variants such as Structural MPI (S-MPI)~\cite{zhang2023structural} and Adaptive MPI~\cite{han2022single} improve local geometry by refining plane orientation or spacing, yet they still lack temporal coherence and boundary stability, especially for large parallax.  
In parallel, neural field–based scene representations (e.g., NeRF and accelerated variants) offer high fidelity but often require heavy sampling and per-scene optimization~\cite{mildenhall2020nerf}.  
Our work stays within the layered family and focuses on content-aligned partitioning with a view-adaptive boundary mechanism (DPS), aiming to retain efficiency while mitigating the classic failure modes.

\subsection{Adaptation to volumetric video.}
Dynamic-scene and volumetric-video representations further extend neural or layered models into the spatio-temporal domain.  
Neural-field approaches such as Neural Volumes~\cite{lombardi2019neural}, D-NeRF~\cite{pumarola2021d}, DyNeRF~\cite{li2022neural}, and 4D Gaussian Splatting~\cite{wu20244d} achieve high-quality 4D reconstruction but rely on multi-view supervision and costly per-scene optimization.  
Hybrid explicit pipelines like V2M4~\cite{chen2025v2m4} provide accurate geometry yet remain storage- and bandwidth-intensive.  
Compact neural-map formulations for volumetric video~\cite{peng2023representing} explore alternative trade-offs between fidelity and speed.  
By contrast, CPSL adapts lightweight layered representations to the volumetric-video setting, keeping layers 2D-encodable while preserving parallax-consistent geometry and low runtime/bitrate, without specialized 4D reconstruction.

\section{Content-Promoted Scene Layer}
\label{sec:cpsl}
Building upon the above motivation, we now introduce the \emph{Content-Promoted Scene Layer (CPSL)}, which serves as the core representation unit of our framework. CPSL extends classical layered 2.5D formats by integrating geometric and perceptual cues to generate compact, content-aware RGBA layers that preserve visual fidelity for immersive scenes, yet remain lightweight and stream-friendly. We first revisit the fundamentals of layered representations and rendering, and then describe the key design components of CPSL—namely, \emph{content-promoted partitioning}, \emph{edge-aware refinement}, and a \emph{view-adaptive synthesizer}. Together, these components enable CPSL to efficiently represent volumetric video with high perceptual fidelity while incurring minimal computational and bandwidth overhead.

\subsection{Preliminaries}
Before detailing the components of CPSL, we first review the fundamentals of layered 2.5D representations and the rendering pipeline on which our method is built. This subsection also introduces the notation used throughout Section 3 and summarizes the geometric assumptions behind layer reprojection and alpha compositing, providing the necessary context for the subsequent modules.

\paragraph{Layered 2.5D Representations.}
A layered 2.5D representation models a scene as a depth-ordered stack of semi-transparent RGBA images, each capturing both appearance and visibility along a reference camera ray~\cite{zhou2018stereo}. 
Formally, a scene is composed of $K$ layers 
$\{L_k\}_{k=1}^{K}$, where each layer 
$L_k = (C_k, \alpha_k, z_k)$ 
consists of a color image 
$C_k \in \mathbb{R}^{H \times W \times 3}$, 
an opacity map 
$\alpha_k \in [0,1]^{H \times W}$, 
and a reference depth value $z_k$. 
The layers are ordered from nearest to farthest 
($z_1 < z_2 < \cdots < z_K$), 
so that occlusion relationships can be handled efficiently during rendering. 
This representation preserves key geometric cues while remaining compatible with standard 2D image operations, forming the foundation of the CPSL framework.

\paragraph{Depth-based Warping.}
Given a layered representation, novel-view synthesis is achieved by geometrically re-projecting each layer according to its depth~\cite{hartley2003multiple}.
Let $p_s = (u_s, v_s, 1)^\top$ and $p_t = (u_t, v_t, 1)^\top$ denote homogeneous pixel coordinates in the source and target cameras, respectively.
For a plane located at depth $z_k$, the mapping between $p_t$ and $p_s$ follows the plane-induced homography:
\begin{equation}
p_s \sim K_s \!\left( R - \tfrac{t\,n^\top}{d} \right)\! K_t^{-1} p_t,
\label{eq:homography}
\end{equation}
where $K_s$ and $K_t$ are the camera intrinsics, $(R,t)$ is the relative rotation and translation from target to source, and $(n,d)$ define the plane normal and distance in the source coordinate system.
For fronto-parallel layers, we set $n = (0,0,1)^\top$ and $d = z_k$.
This formulation enables accurate reprojection of each depth slice without explicit 3D reconstruction, serving as the geometric basis for view synthesis in CPSL.

\paragraph{Alpha Compositing.}
After geometric reprojection, the final novel view is synthesized through front-to-back alpha blending, given the color and opacity maps $\{C_k, \alpha_k\}_{k=1}^{K}$, the pixel intensity at $(u,v)$ is computed as:
\begin{equation}
I(u,v) = \sum_{k=1}^{K} C_k(u,v) \, \alpha_k(u,v) \prod_{j=1}^{k-1} \big(1 - \alpha_j(u,v)\big),
\label{eq:alpha}
\end{equation}
where the product term accumulates the transmittance of all preceding layers.
This formulation ensures that nearer layers naturally occlude farther ones while preserving semi-transparent boundaries.
In practice, premultiplied color is used to avoid fringe artifacts during blending, yielding visually smooth transitions across depth discontinuities~\cite{porter1984compositing}.





\begin{figure*}[t]
\centering
\includegraphics[width=\textwidth]{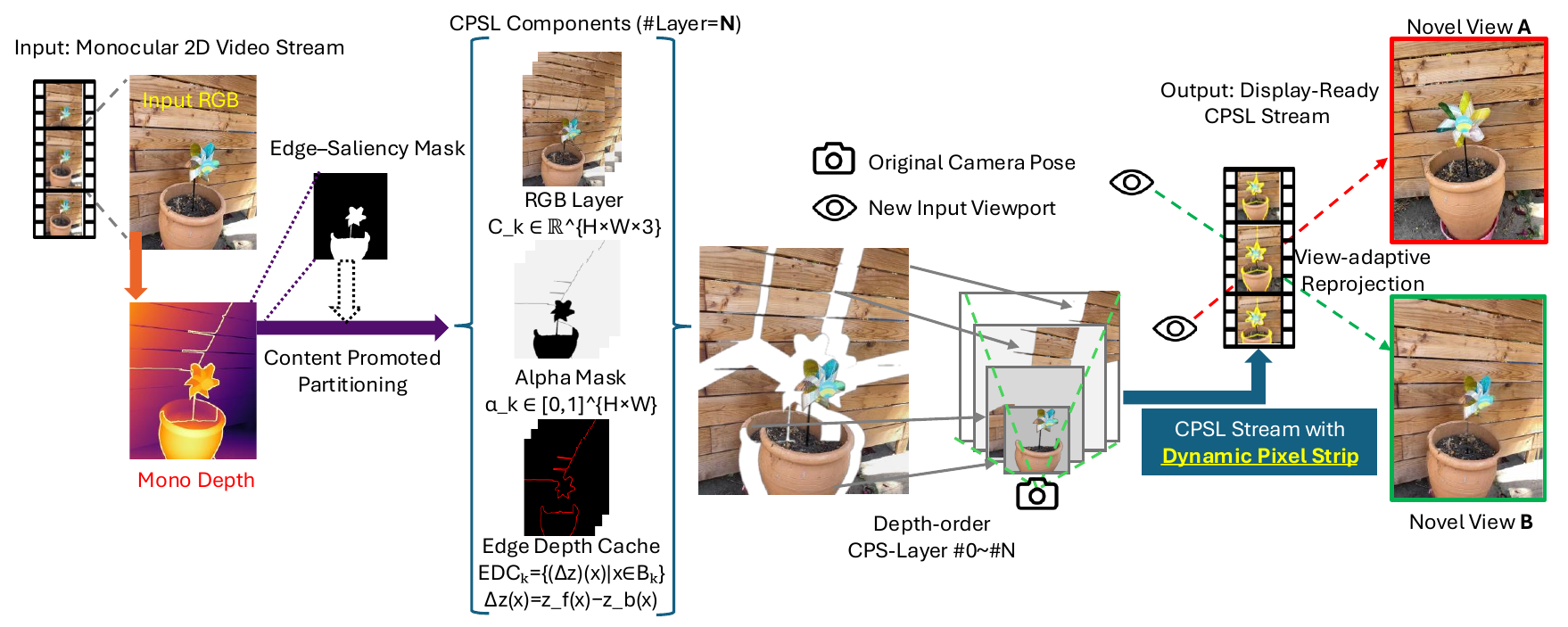}
\caption{
\textbf{Overview of CPSL.}
CPSL converts a monocular video into a compact set of content-promoted scene layers that preserve parallax and boundary integrity. 
Each frame is decomposed into depth-ordered RGBA layers guided by depth and semantic cues, producing geometry-consistent and temporally coherent assets. 
For video deployment, these layers are propagated across frames in a lightweight GOP structure and encoded as 2D streams with edge–depth cache (EDC). 
During playback, a Dynamic Pixel Strip (DPS) maintains continuous silhouettes and crack-free parallax, enabling real-time volumetric rendering with spatial fidelity.
}

\label{fig:pipeline_layer_generation}
\end{figure*}

\subsection{Layer-Set Generation}
\label{sec:layer-generation}
Building on the above preliminaries, we now describe how
each input video frame is decomposed into a compact,
depth-ordered collection of semi-transparent RGBA layers
that constitute the CPSL representation. The objective is to preserve sharp object boundaries and accurate occlusion ordering while maintaining compactness for efficient encoding. Unlike purely depth-based slicing, which often produces noisy silhouettes and fragmented regions\cite{shih20203d}, CPSL integrates geometric (depth) and semantic cues to generate object-aligned, perceptually consistent layers suitable for immersive rendering.

\paragraph{Depth–semantic Fusion.}
Depth-only slicing often yields jagged boundaries and mixed-object fragments near silhouettes~\cite{shih20203d}.
To obtain perceptually coherent layers, we formulate layer assignment as a content-aware energy minimization that jointly considers depth reliability, semantic saliency, and spatial smoothness:
\begin{equation}
\label{eq:cpsl_energy}
\begin{aligned}
\mathcal{E}(S) &=
\sum_{x}
\Big[
w_d(x)\, \rho\!\big(z(x) - \bar{z}_{S(x)}\big)
+ w_s(x)\, \phi\!\big(c(x), S(x)\big)
\\[-3pt]
&\hspace{-0.8em}
+\, w_i(x)\, \psi\!\big(\iota(x), S(x)\big)
\Big]
+ \lambda_b
\!\!\!\!\sum_{(x,y)\in\mathcal{N}}\!\!\!\!
\omega_{xy}\, \mathbf{1}[S(x)\neq S(y)] .
\end{aligned}
\end{equation}
where $S(x)$ denotes the assigned layer index for pixel $x$, 
$z(x)$ and $\bar{z}_{S(x)}$ are the observed and representative depths, 
and $\rho(\cdot)$ is a robust penalty.
The weights $w_d$, $w_s$, and $w_i$ encode pixel-wise confidence:
\begin{itemize}
\item $w_d(x)=\varsigma_z(x)$ measures depth stability, 
\item $w_s(x)=\sigma(\text{saliency}(x))$ emphasizes visually important regions, and 
\item $w_i(x)$ enforces instance-level coherence.
\end{itemize}
The pairwise term uses
$\omega_{xy} = \exp[-(\alpha\|\nabla I(x,y)\| + \beta b_{\text{sem}}(x,y))]$ 
to encourage smoothness while preserving semantic edges $b_{\text{sem}}$.
Minimizing $\mathcal{E}$ yields a depth and content consistent layer map
that partitions the scene into compact, object-aligned regions.

\paragraph{Instance Promotion and Background Merging.}
Given the preliminary pixel-wise layer map $S(x)$ from depth–semantic fusion,
we further group pixels into coherent instances and aggregate them into a compact
layer set. Salient or foreground instances (e.g., people, hands, or tools) are
\emph{promoted} to dedicated layers to preserve silhouettes and occlusion ordering,
while non-salient background regions are clustered by depth and texture statistics
and subsequently \emph{merged} to satisfy a target layer count $K$ and bitrate budget.
This stage balances perceptual fidelity with compactness: salient regions maintain
structural integrity, whereas background areas are consolidated based on low depth
variance and weak semantic gradients. Promotion is guided by a joint saliency score
derived from the fused saliency and semantic confidence maps, integrating instance
size and local visual importance to ensure that each resulting layer corresponds to
a semantically meaningful, geometrically stable scene element.

\paragraph{Layer Matting and Temporal Coherence.}
To ensure spatially smooth and temporally stable layer boundaries,
we refine each layer’s alpha matte and maintain consistent instance tracking across frames.
For each layer $k$, a soft matte $\alpha_k(x)$ is generated from the signed distance to its semantic contour, with adaptive feather width $w = w_0 + a\|\nabla z\| + b\varsigma_z$ to accommodate depth gradients and uncertainty. To prevent dark-edge artifacts introduced by standard non-premultiplied blending, we adopt premultiplied colors where the RGB channels are linearly scaled by the alpha matte~\cite{porter1984compositing}. This formulation maintains correct color accumulation across semi-transparent feather regions and eliminates halo artifacts at depth discontinuities. Across frames, instances are associated using optical flow~\cite{teed2020raft} and mask IoU (Intersection-over-Union), allowing layer indices and alpha mattes to be propagated over time. Exponential moving averages are applied near boundaries, where semantic edges dominate updates under unstable depth estimates. These refinements ensure geometry-aligned and temporally coherent CPSL layers for robust rendering.

\begin{figure}[t]
\centering

\begin{minipage}[c]{0.48\linewidth}
    \centering
    \includegraphics[width=\linewidth]{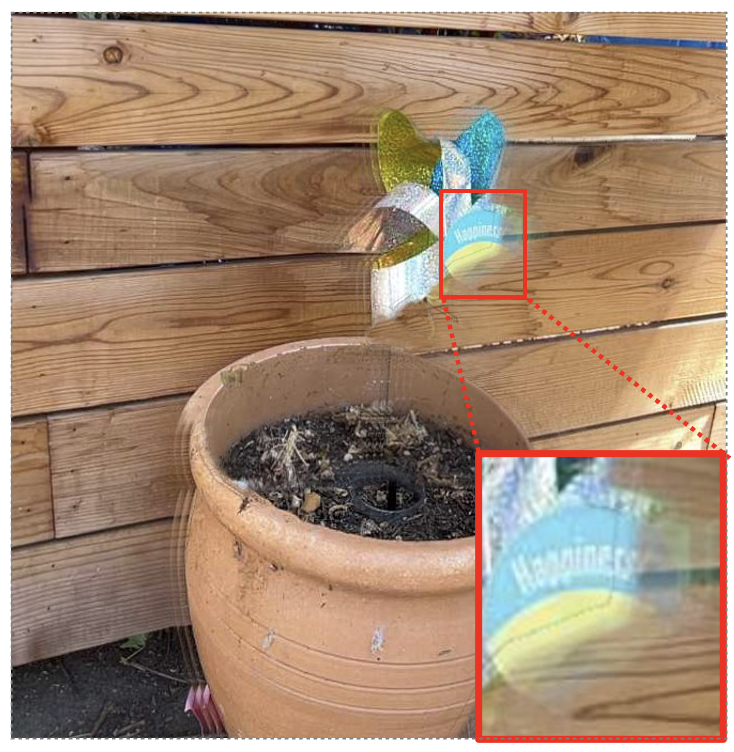}
    \subcaption*{\textbf{Without DPS}}
\end{minipage}
\hfill
\begin{minipage}[c]{0.48\linewidth}
    \centering
    \includegraphics[width=\linewidth]{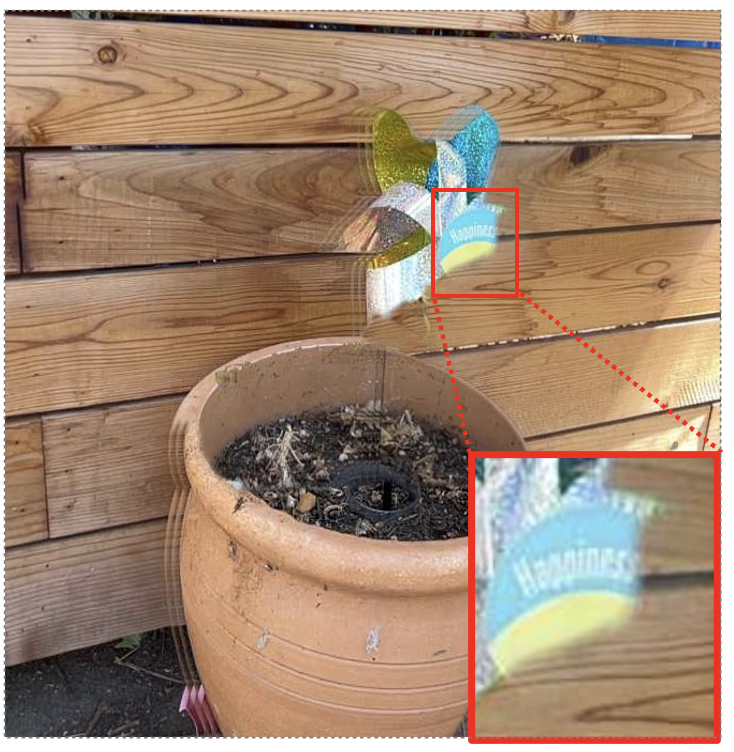}
    \subcaption*{\textbf{With DPS}}
\end{minipage}

\caption{\textbf{Dynamic Pixel Strip (DPS) comparison}. 
Without DPS, direct per-layer warping causes cracks and double edges near depth discontinuities. With DPS, a lightweight view-adaptive transition strip restores boundary continuity and produces clean silhouettes under viewpoint motion.}
\vspace{-10pt}
\label{fig:dps}
\end{figure}

\subsection{Dynamic Pixel Strip}
\label{sec: dyanmic pixel strip}
Layered 2.5D renderings often exhibit cracks, halos, and double edges near depth discontinuities when reprojected to novel viewpoints~\cite{zhou2018stereo,shih20203d,tucker2020single,choi2019extreme}. These artifacts arise because neighboring layers are warped independently, causing small inconsistencies in depth or alpha to manifest as visible gaps around high-contrast silhouettes. Such discontinuities disrupt geometric coherence and become increasingly distracting under viewpoint motion.

To mitigate these effects, we introduce the \emph{Dynamic Pixel Strip} (DPS), a lightweight screen-space transition band constructed \emph{after} all CPSL layers are warped into the viewer’s frame. DPS reconnects foreground and background along detected silhouette boundaries by synthesizing a narrow strip of intermediate pixels whose appearance and depth smoothly interpolate between the warped layers. This preserves edge continuity without modifying the original assets or requiring full-frame depth maps.

Formally, let $L_f=(C_f,\alpha_f,z_f)$ and $L_b=(C_b,\alpha_b,z_b)$ denote the warped foreground and background layers with $z_f < z_b$. Along a boundary band $\mathcal{B}$, DPS generates interpolated values:
\begin{equation}
\begin{aligned}
z_{\mathrm{dps}}(x) &= (1-\gamma(x))\, z_f(x) + \gamma(x)\, z_b(x),\\
C_{\mathrm{dps}}(x) &= (1-\gamma(x))\, C_f(x) + \gamma(x)\, C_b(x),
\end{aligned}
\label{eq:dps_blend}
\end{equation}
where $\gamma(x)\in[0,1]$ is an adaptive blend weight modulated by local parallax and the geometry of the projected silhouette. This interpolation bridges layer misalignments in screen space while preserving correct depth ordering.

In practice, DPS identifies silhouette regions using warped alpha gradients and constructs a thin strip aligned to these boundaries. The strip’s support and blend weights are updated each frame based on the viewer’s pose, yielding spatially continuous and temporally stable edges even under large viewpoint shifts. To avoid storing dense depth maps, DPS relies on a boundary-only \emph{Edge-Depth Cache} (EDC), which provides accurate local depth differences for robust, geometry-aware blending. Additional implementation and comparison details are provided in supplementary material.

\subsection{View-Adaptive Layer Reprojection and Compositing}

The view-adaptive synthesizer reconstructs novel views by reprojecting and compositing the CPSL layers based on the viewer’s current viewpoint. Unlike neural rendering pipelines that trace rays through continuous 3D fields ~\cite{mildenhall2020nerf,muller2022instant,yariv2021volume}, CPSL enables real-time view synthesis using only plane-induced homographies and alpha blending. This makes the rendering stage lightweight and well suited for mobile or streaming environments while preserving accurate geometric parallax.

\paragraph{Virtual Parallax.}
Given the viewer’s pose, expressed as the relative rotation–translation $(R, t)$ between the reference and target cameras, each CPSL layer is warped into the viewer’s screen space. The resulting virtual parallax represents the apparent pixel displacement of scene content caused by the viewpoint shift, and serves as the geometric basis for generating a consistent novel view.

\paragraph{Layer Reprojection.}
For a pixel $(x, y)$ on layer $k$ with depth $z_k$, the corresponding camera-space point is expressed as
\[
X_s = z_k K_s^{-1} [x, y, 1]^\top.
\]

The point is transformed to the viewer’s camera by $X_t = R X_s + t$ and projected via
\[
p_t \sim K_t X_t,
\]
yielding the pixel’s new location in the synthesized view. This homography-based reprojection is applied independently to the per-layer attributes of each CPSL layer. Since layer depths encode fronto-parallel slices, this transformation produces accurate parallax without requiring full 3D reconstruction.

\paragraph{Alpha Compositing.}
Once reprojected, layers are composited in front-to-back order using premultiplied alpha:
\begin{equation}
I(u,v) = \sum_{k=1}^{K} C_k(u,v)\,\alpha_k(u,v)
\prod_{j=1}^{k-1} (1 - \alpha_j(u,v)).
\end{equation}

This blends semi-transparent feather regions while ensuring that nearer layers occlude farther ones. Combined with the Dynamic Pixel Strip (Sec.~\ref{sec: dyanmic pixel strip}), the procedure yields silhouette-consistent, crack-free boundaries under large viewpoint changes.

\paragraph{Practical Considerations.}
The viewer’s pose is treated as an external input obtained 
from standard head tracking, device motion, or cursor-driven 
controls. The synthesizer’s computation is $O(KHW)$ with 
inexpensive 2D warps, enabling $>$60\,FPS real-time rendering 
on commodity GPUs. Together, these components form an efficient 
pipeline that converts CPSL assets into stable, parallax-corrected 
novel-view video frames.



\begin{figure*}[t]
\centering
\setlength{\tabcolsep}{1pt}
\renewcommand{\arraystretch}{0.96}


\begin{tabular}{L{0.05\textwidth} *{6}{C{0.140\textwidth}}}
    & \textbf{$0^\circ$} & \textbf{$5^\circ$} & \textbf{$10^\circ$} &
      \textbf{$15^\circ$} & \textbf{$20^\circ$} & \textbf{$30^\circ$} \\
    \rotatebox{90}{\textbf{PtCl}} &
    \includegraphics[height=0.115\textwidth]{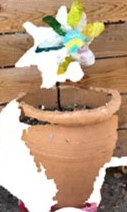} &
    \includegraphics[height=0.115\textwidth]{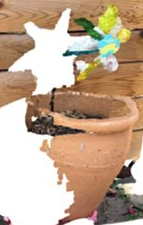} &
    \includegraphics[height=0.115\textwidth]{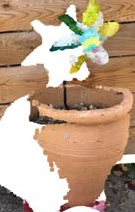} &
    \includegraphics[height=0.115\textwidth]{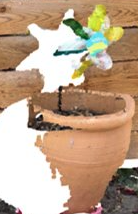} &
    \includegraphics[height=0.115\textwidth]{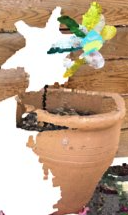} &
    \includegraphics[height=0.115\textwidth]{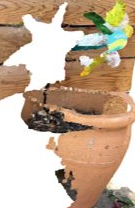} \\
    \rotatebox{90}{\textbf{MPI}} &
    \includegraphics[height=0.115\textwidth]{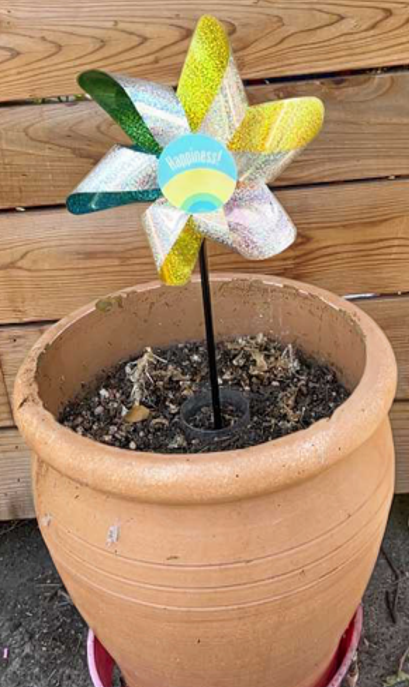} &
    \includegraphics[height=0.115\textwidth]{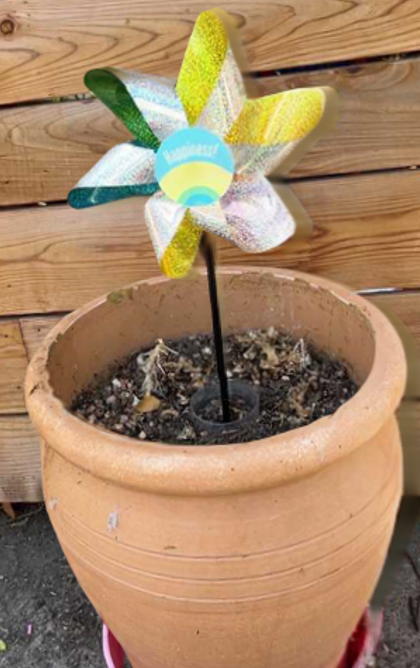} &
    \includegraphics[height=0.115\textwidth]{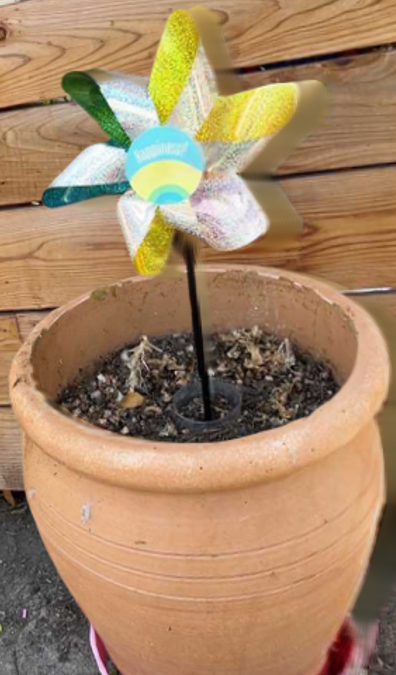} &
    \includegraphics[height=0.115\textwidth]{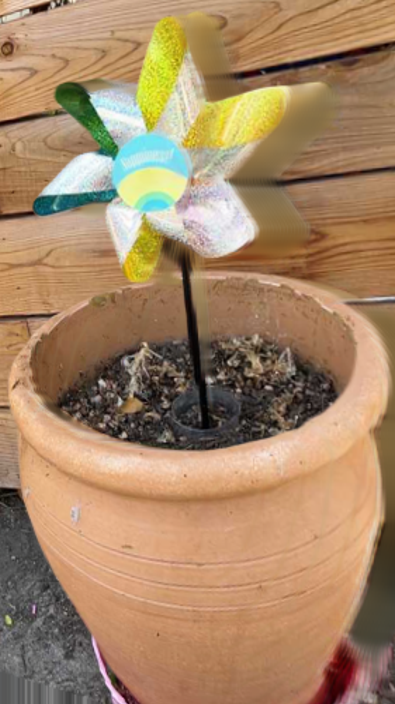} &
    \includegraphics[height=0.115\textwidth]{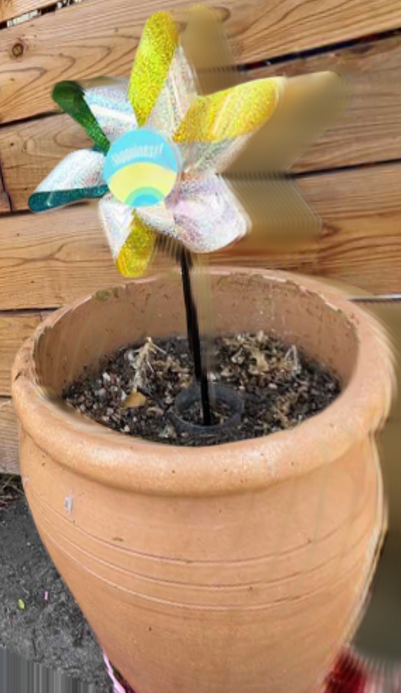} &
    \includegraphics[height=0.115\textwidth]{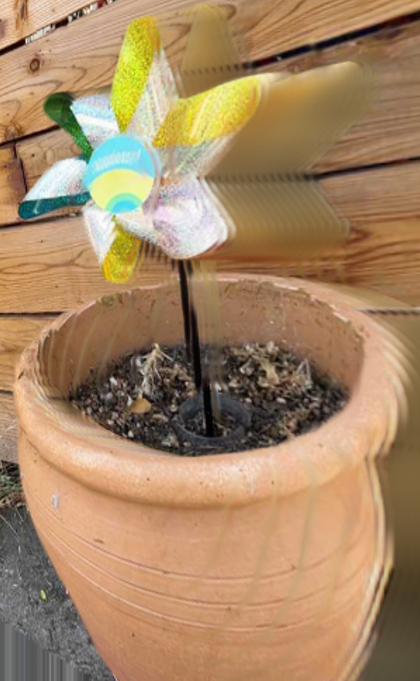} \\
    \rotatebox{90}{\textbf{4DGS}} &
    \includegraphics[height=0.115\textwidth]{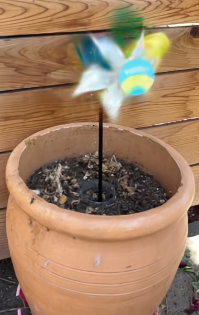} &
    \includegraphics[height=0.115\textwidth]{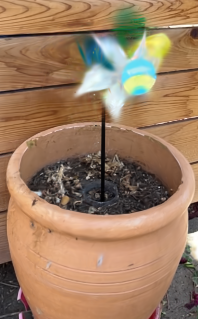} &
    \includegraphics[height=0.115\textwidth]{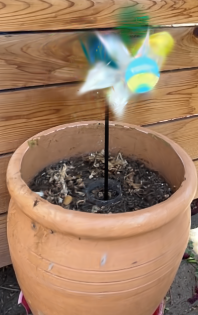} &
    \includegraphics[height=0.115\textwidth]{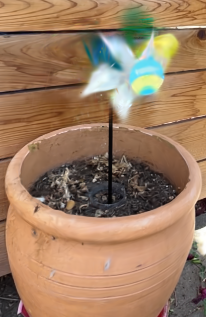} &
    \includegraphics[height=0.115\textwidth]{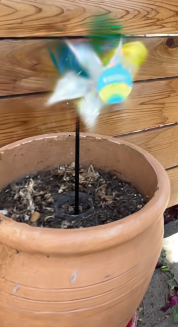} &
    \includegraphics[height=0.115\textwidth]{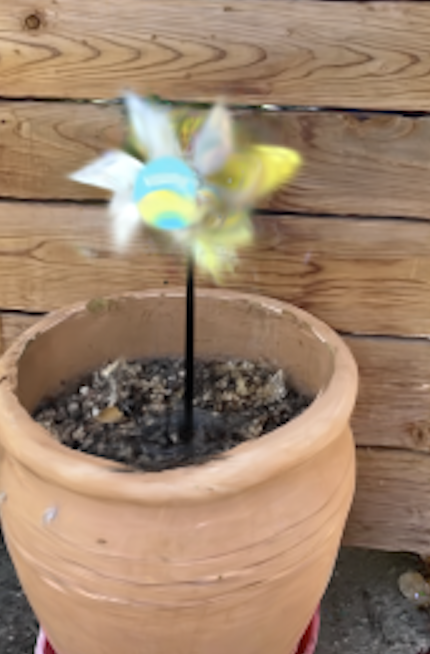} \\
    \rotatebox{90}{\textbf{CPSL(ours)}} &
    \includegraphics[height=0.115\textwidth]{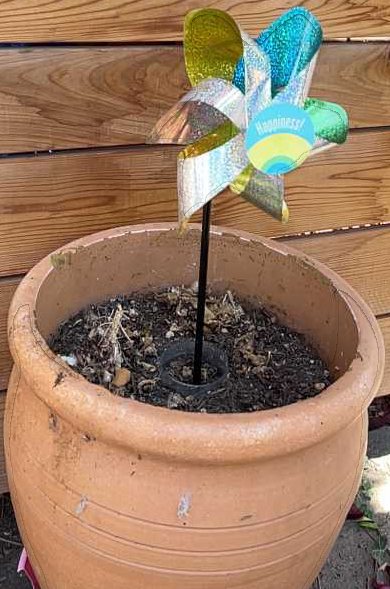} &
    \includegraphics[height=0.115\textwidth]{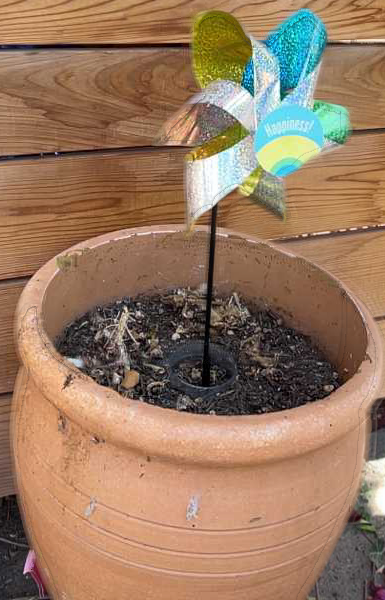} &
    \includegraphics[height=0.115\textwidth]{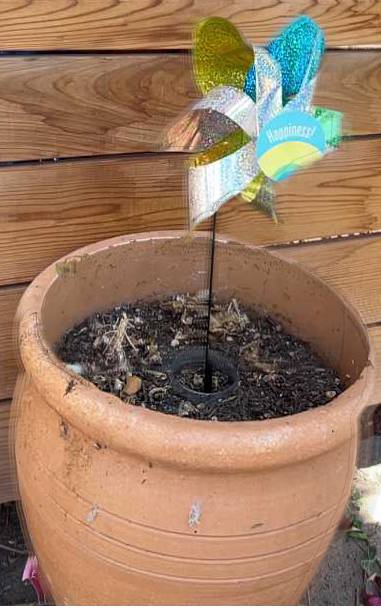} &
    \includegraphics[height=0.115\textwidth]{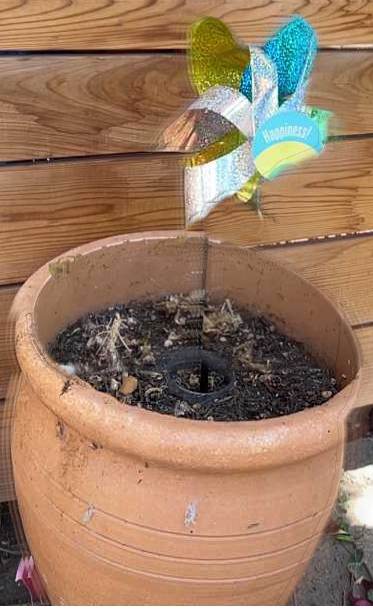} &
    \includegraphics[height=0.115\textwidth]{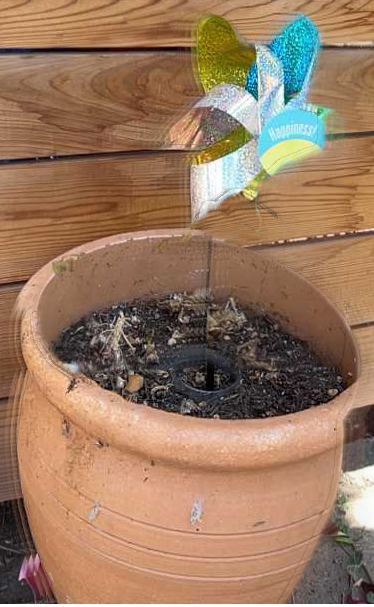} &
    \includegraphics[height=0.115\textwidth]{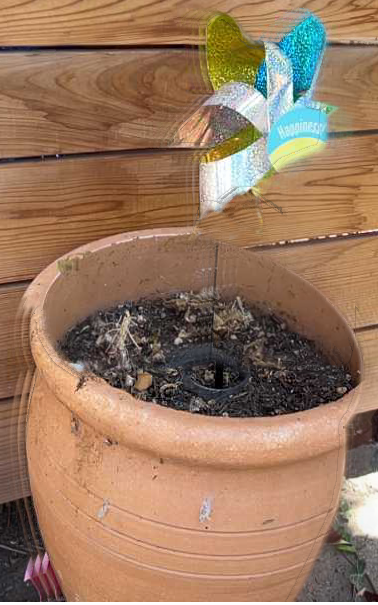} \\
\end{tabular}

\vspace{-0.6em}
\caption{
\textbf{Multi-method viewpoint sweep (0°–30°).}
Each row corresponds to a different representation baseline, and each column shows the novel-view rendering at a specific viewpoint offset.
CPSL preserves sharp boundaries and parallax-consistent geometry across large viewpoint changes, whereas depth-only and layered baselines produce cracks or double edges, and neural-field methods exhibit smoothing or ghosting. \textbf{Note} that 4DGS receives multi-frame video supervision, whereas CPSL and other 2.5D baselines operate strictly from single-view monocular inputs.}
\label{fig:view_sweep_matrix}
\vspace{-0.8em}
\end{figure*}

\section{2.5D Video Representation via CPSL}
\label{sec:video_cpsl}

While Section~\ref{sec:cpsl} establishes CPSL as a compact, content-aware layered representation for individual frames, deploying CPSL in video settings introduces additional challenges. This section extends CPSL from a frame-wise representation to a temporally coherent and stream-friendly 2.5D video format that supports parallax-corrected novel-view synthesis without requiring multi-view reconstruction or neural field training. An overview of the full pipeline is shown in Fig.~\ref{fig:pipeline_layer_generation}.

\subsection{Temporal Organization}

A CPSL video is organized as a sequence of layer sets  $\{\mathcal{L}_t\}$ that preserve both semantic and geometric consistency over time.
We adopt a lightweight \textit{GOP} structure~\cite{wiegand2003overview} to balance stability and efficiency: 
key frames (\emph{I-frames}) are fully decomposed into $K$ content-promoted layers, whereas intermediate (\emph{P-frames}) reuse and update the previous layers through decoder-provided \textit{motion-vector propagation}. For each layer $L^k_t$, a temporal confidence value $c^k_t$ is maintained by an exponential moving average of mask overlap and boundary similarity. When the average intersection-over-union (IoU) between successive masks drops below a threshold or the local crack rate exceeds another, the system triggers an automatic I-frame refresh. This mechanism ensures that layer semantics remain stable while updates are localized to regions of real structural change, making per-frame complexity sublinear in the layer count $K$. The result is a temporally smooth sequence of content-promoted layers that evolve coherently with scene motion.

\subsection{Encoding and Packaging}

CPSL treats layers as the fundamental perceptual units of
coding.  
Each layer is encoded as an independent RGBA stream using
standard H.265/AV1 codecs~\cite{sze2014high}, while edge-depth cache (\emph{EDC}) and temporal
confidence are recorded in lightweight side channels.
The EDC stores sparse, quantized depth offsets along layer
boundaries and acts as a compact geometric prior that
complements the $\alpha$-band, allowing boundary integrity
to be restored at playback with negligible cost.
Given a total bitrate budget $R$, per-layer bitrates
$\{r_k\}$ are allocated by minimizing a saliency and
boundary-weighted perceptual cost:
\begin{equation}
\min_{\{r_k\}} \sum_k w_k\,\mathrm{LPIPS}_k(r_k)
\quad \text{s.t.}\quad \sum_k r_k \le R ,
\label{eq:alloc}
\end{equation}
where $w_k$ increases with visual saliency and depth-edge
sharpness, and $\mathrm{LPIPS}_k(r_k)$ monotonically
decreases with $r_k$.
Foreground or edge-rich layers are thus prioritized, whereas
stable backgrounds may be merged or down-weighted under
bandwidth constraints.

\subsection{Runtime Playback}

At playback, decoded layers are warped to the target camera
pose via plane-induced homographies and composited in
back-to-front order using premultiplied $\alpha$-blending:
\[
C_{\text{out}} = \sum_k \alpha_k C_k
\prod_{j<k} (1 - \alpha_j).
\]

Residual misalignments are then corrected in screen space using a
lightweight boundary refinement step guided by the cached
EDC samples. This refinement operates in real time and
preserves boundary sharpness without introducing temporal
flicker.

\begin{table*}[t]
\centering
\small
\setlength{\tabcolsep}{8pt}
\renewcommand{\arraystretch}{1.05}

\begin{tabular}{l l l c c c c}
\toprule
\textbf{Method} & \textbf{Input} & \textbf{Type} & \textbf{LPIPS↓} & \textbf{PSNR↑} & \textbf{SSIM↑} & \textbf{Crack↓} \\
\midrule
DIBR~\cite{artois2024opendibr}            & Mono-frame  & Depth/2.5D & 0.24 & 24.70 & 0.82 & 0.14 \\
MPI~\cite{zhang2023structural}             & Mono-frame  & 2.5D       & 0.18 & 26.50 & 0.87 & 0.11 \\
4D-GS (Shape of Motion)~\cite{wang2025shape} & Mono-video  & 4D         & 0.12 & 28.80 & 0.92 & — \\
V2M4 (4D Mesh)~\cite{chen2025v2m4}         & Mono-video  & Mesh       & 0.20 & 25.20 & 0.84 & — \\
Depth2PC~\cite{de2025low}                  & Mono-frame  & Depth/PC   & 0.21 & 25.30 & 0.85 & — \\
\textbf{CPSL (ours)}                       & Mono-frame  & 2.5D       & \textbf{0.11} & \textbf{29.60} & \textbf{0.94} & \textbf{0.05} \\
\bottomrule
\end{tabular}
\caption{
\textbf{DyCheck monocular novel-view evaluation (0°/5°/10°/15°/20°/30°).}
Quantitative comparison on DyCheck using monocular RGB input.
Metrics are averaged over sequences and co-visible pixels; Crack is computed on boundary bands.
`--' indicates the metric is not applicable or not reported.
Note that 4DGS (Shape of Motion) and V2M4 (4D Mesh) are multi-frame video-based methods,
whereas CPSL and other 2.5D approaches operate strictly from monocular frames.
CPSL achieves the best perceptual quality and boundary stability among 2.5D methods,
outperforming MPI by +3.1\,dB PSNR and reducing Crack by about 2$\times$.
}
\label{tab:dycheck_main}
\vspace{-0.6em}
\end{table*}

\section{Evaluation}
\label{sec:evaluation}

We evaluate CPSL across three dimensions essential for
monocular 2.5D video representations: 
(1) per-frame spatial fidelity, 
(2) novel-view rendering quality, and 
(3) temporal stability under camera and scene motion.
Across all experiments, CPSL is compared against representative depth-based 2.5D methods, and to neural scene representations on sequences where monocular inputs are supported.

\subsection{Experiment Setup}
\label{sec:exp_setup}

\paragraph{Datasets.}
We evaluate CPSL on the \textbf{DyCheck} benchmark~\cite{gao2022monocular},
which features casual handheld iPhone captures of real-world dynamic scenes. DyCheck provides calibrated frames, co-visibility masks, and diverse motion patterns that closely match our monocular target setting. Since CPSL operates strictly with single-camera inputs, DyCheck naturally aligns with our evaluation protocol.
All compared methods use the same monocular video inputs to ensure consistency.

\paragraph{Baselines.}
We compare CPSL with the monocular-capable baselines reported in Table~\ref{tab:dycheck_main}, covering both classical 2.5D warping and dynamic-scene reconstruction methods. Specifically, we include: (1) DIBR~\cite{artois2024opendibr}, a per-frame depth-warping baseline widely used for single-view novel-view synthesis; (2) MPI~\cite{zhou2018stereo}, a layered 2.5D representation based on fronto-parallel multi-plane decomposition; (3) 4D Gaussian Splatting (4D-GS)~\cite{wu20244d}, reported as an upper-bound dynamic-scene reference, though it typically requires multi-frame supervision; (4) V2M4~\cite{chen2025v2m4}, a monocular 4D mesh reconstruction method; and (5) Depth2PC ~\cite{zhong2024low}, which converts monocular depth into point clouds for direct geometric rendering. All baselines are evaluated using the same monocular input frames to ensure consistency and fairness.

\paragraph{Metrics.}
We evaluate novel-view fidelity using PSNR, SSIM, and LPIPS, and measure boundary accuracy using Crack on co-visible regions 
following DyCheck~\cite{gao2022monocular}. 
Temporal stability is assessed with boundary variance, F-SSIM, and flicker score.


\subsection{Visual Quality}
\label{sec:visual_quality}

Figure~\ref{fig:view_sweep_matrix} compares viewpoint sweeps (0°–30°) 
across all baselines. Each row shows a different representation, and 
each column shows the novel view at a specific offset. The analysis on visual behavior along three aspects is reflected.

\paragraph{Novel-View Fidelity.}
Across viewpoint sweeps (0°–30°), CPSL delivers consistently higher novel-view fidelity than depth-only warping and MPI. By enforcing per-layer geometric consistency and boundary-aware compositing, CPSL preserves thin structures, high-frequency textures, and silhouette detail. The improvements are most visible in regions with strong parallax and intricate silhouettes, where other 2.5D methods frequently suffer from distortion, boundary artifacts, or missing geometry.

\paragraph{Disocclusion Handling.}
Dynamic scenes in DyCheck contain frequent occlusion–disocclusion transitions arising from hand motion, articulation, and camera shake. These events reveal misalignments in depth-only and layered baselines, often leading to cracks, double edges, and floating pixels. CPSL mitigates these artifacts with Dynamic Pixel Strip (DPS), which adaptively interpolates foreground and background within silhouette-adjacent bands. This maintains clean boundaries and coherent disocclusion regions even under large viewpoint offsets (e.g., $>15^\circ$).

\paragraph{Parallax Accuracy.}
CPSL achieves stable and geometrically consistent parallax across layers due to its fronto-parallel per-layer depths and content-promoted layer partitioning. Under large viewpoint shifts, MPI tends to over-smooth slanted surfaces, while Gaussian-splatting methods blur or smear fine-scale texture. In contrast, CPSL preserves both structure and detail over the full viewpoint sweep, demonstrating strong robustness to long-range parallax motion on DyCheck.


\begin{table}[t]
\centering

\resizebox{\columnwidth}{!}{%
\begin{tabular}{lccc}
\toprule
Method & Boundary Var.$\downarrow$ & F-SSIM$\uparrow$ & Flicker Score$\downarrow$ \\
\midrule
Frame-wise Depth (baseline) & 1.00 & 0.910 & 0.128 \\
I/P Propagation (no refine) & 0.63 & 0.935 & 0.082 \\
I/P + Refinement + Hysteresis & \textbf{0.47} & \textbf{0.948} & \textbf{0.059} \\
\bottomrule
\end{tabular}
}
\caption{Ablation on temporal stability for CPSL playback.
Metrics: lower variance and higher F-SSIM indicate smoother video.}
\label{tab:temporal}
\end{table}

\subsection{Temporal Stability and Flicker Reduction}

We assess temporal consistency under both fixed and moving viewpoints. 
Depth-based 2.5D renderings typically suffer from silhouette flicker and unstable layer boundaries because each frame is warped independently. 
CPSL reduces these temporal artifacts through geometry-aligned soft matting, GOP-based layer propagation, and the Dynamic Pixel Strip (DPS), which jointly maintain coherent boundary transitions across time.

\paragraph{Quantitative Metrics.}
Table~\ref{tab:temporal} evaluates temporal stability using boundary variance, F-SSIM, and flicker score. 
Frame-wise depth warping yields the largest boundary variance and highest flicker, reflecting inconsistent silhouette alignment across frames. 
Introducing cross-frame propagation substantially stabilizes layer evolution, and the complete CPSL pipeline achieves the lowest variance and flicker while also attaining the highest F-SSIM, indicating smooth temporal transitions without degrading spatial detail.

\paragraph{Temporal Stability Analysis.}
The trends in Table~\ref{tab:temporal} are consistent with the expected behavior of different representations. 
Directly reconstructing depth independently for each frame cannot enforce coherent layer boundaries, causing temporal jitter and abrupt texture changes, captured by higher boundary variance and elevated flicker scores. 
In contrast, full CPSL pipeline maintains stable silhouettes by propagating content layers over consecutive frames and refining boundary alignment, leading to markedly smoother playback and significantly reduced flicker.

\begin{table}[t]
\centering

\resizebox{\columnwidth}{!}{%
\begin{tabular}{lcccc}
\toprule
Representation & PSNR↑ & SSIM↑ & LPIPS↓ & Bitrate↓ (Mbps) \\
\midrule
Depth-based Point Cloud~\cite{zhong2024low} & 29.4 & 0.932 & 0.118 & 18.2 \\
Adaptive MPI~\cite{luvizon2021adaptive}                          & 30.1 & 0.945 & 0.105 & 9.6 \\
CPSL (Ours)                                   & \textbf{31.2} & \textbf{0.956} & \textbf{0.083} & \textbf{2.3} \\
\bottomrule
\end{tabular}
}
\caption{
\textbf{User-centric VV playback on FSVVD.}
CPSL is compared with depth-based point cloud streaming~\cite{zhong2024low}
and Adaptive MPI.
Results are measured at 1080p and 60\,FPS playback.
CPSL achieves similar perceptual quality while reducing bitrate by $7\times$.
Additional quantitative and visual results are provided in the supplementary material.
}
\label{tab:fsvvd}
\end{table}

\begin{figure}[t]
\centering
\setlength{\tabcolsep}{2pt}

\begin{tabular}{cc}
    \includegraphics[width=0.47\linewidth]{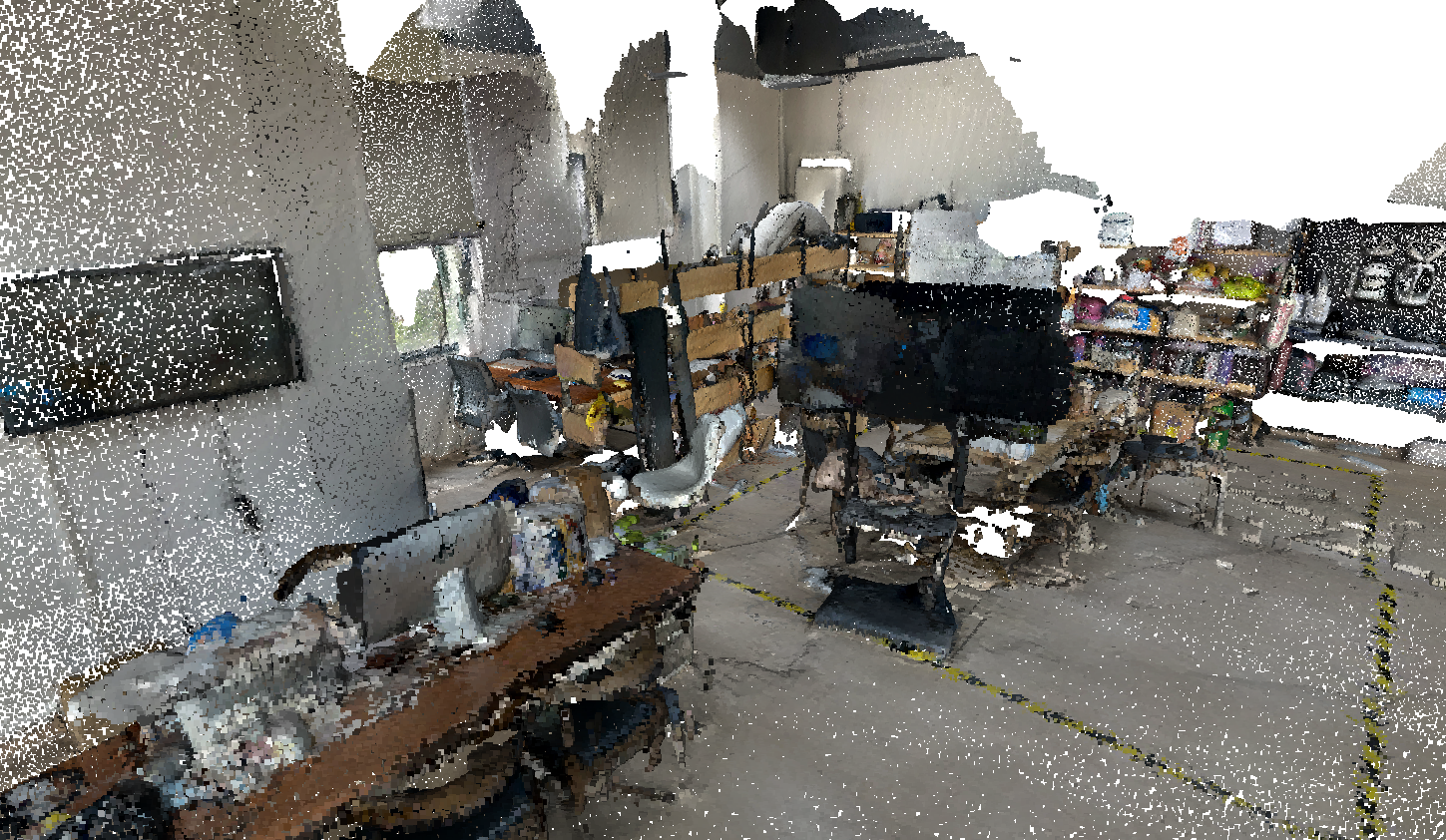} &
    \includegraphics[width=0.47\linewidth]{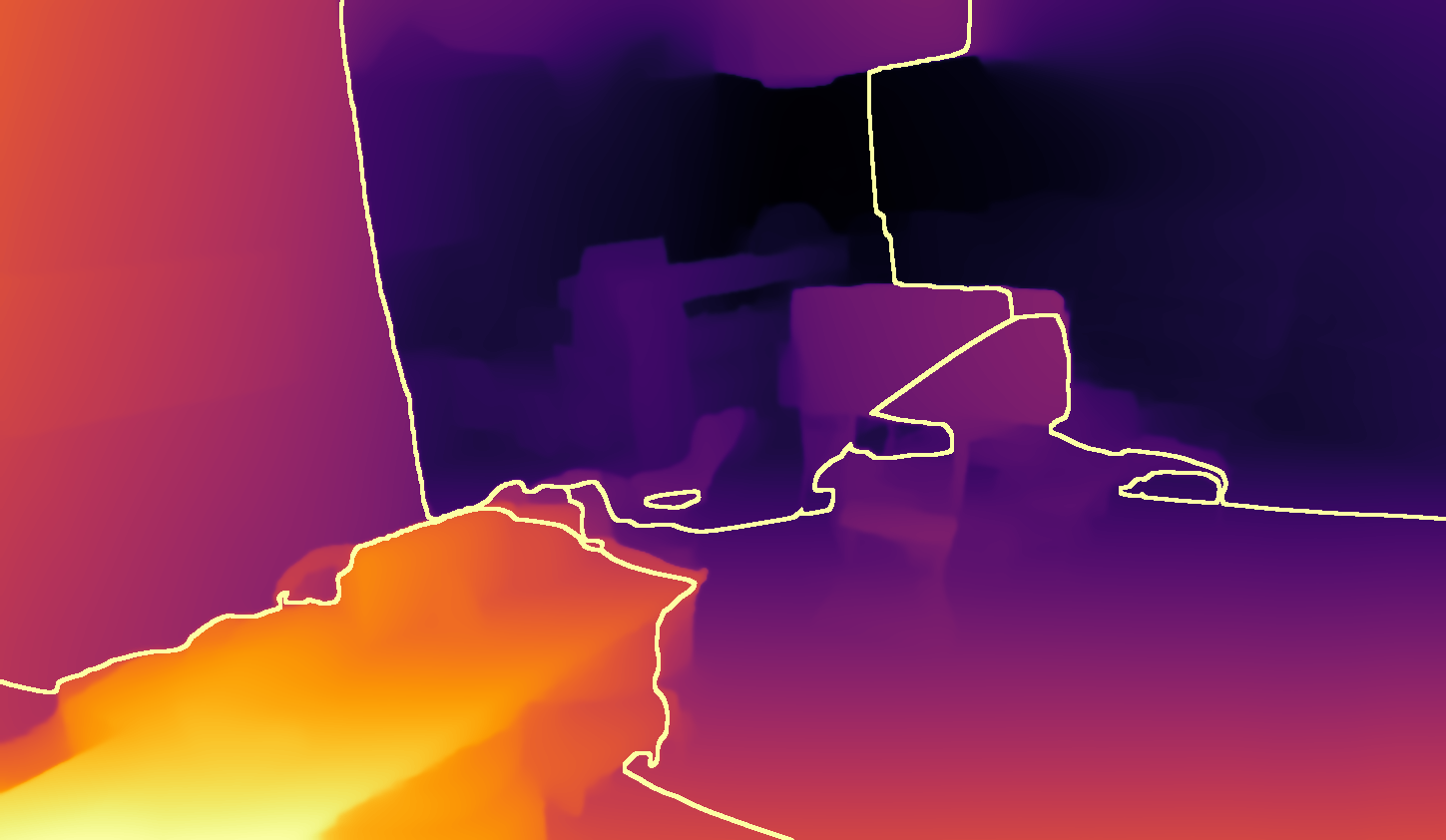} \\
    {\small (a) Original FSVVD Rendering} &
    {\small (b) CPSL Depth} \\
    
    \includegraphics[width=0.47\linewidth]{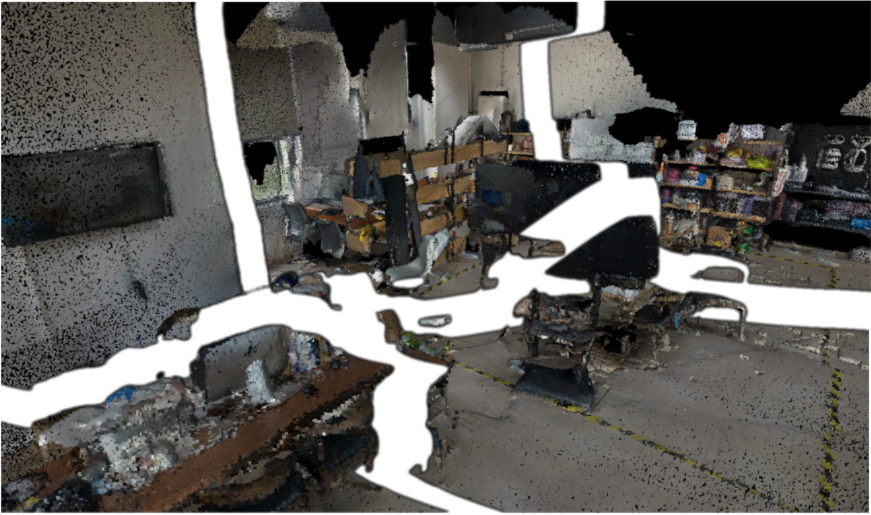} &
    \includegraphics[width=0.47\linewidth]{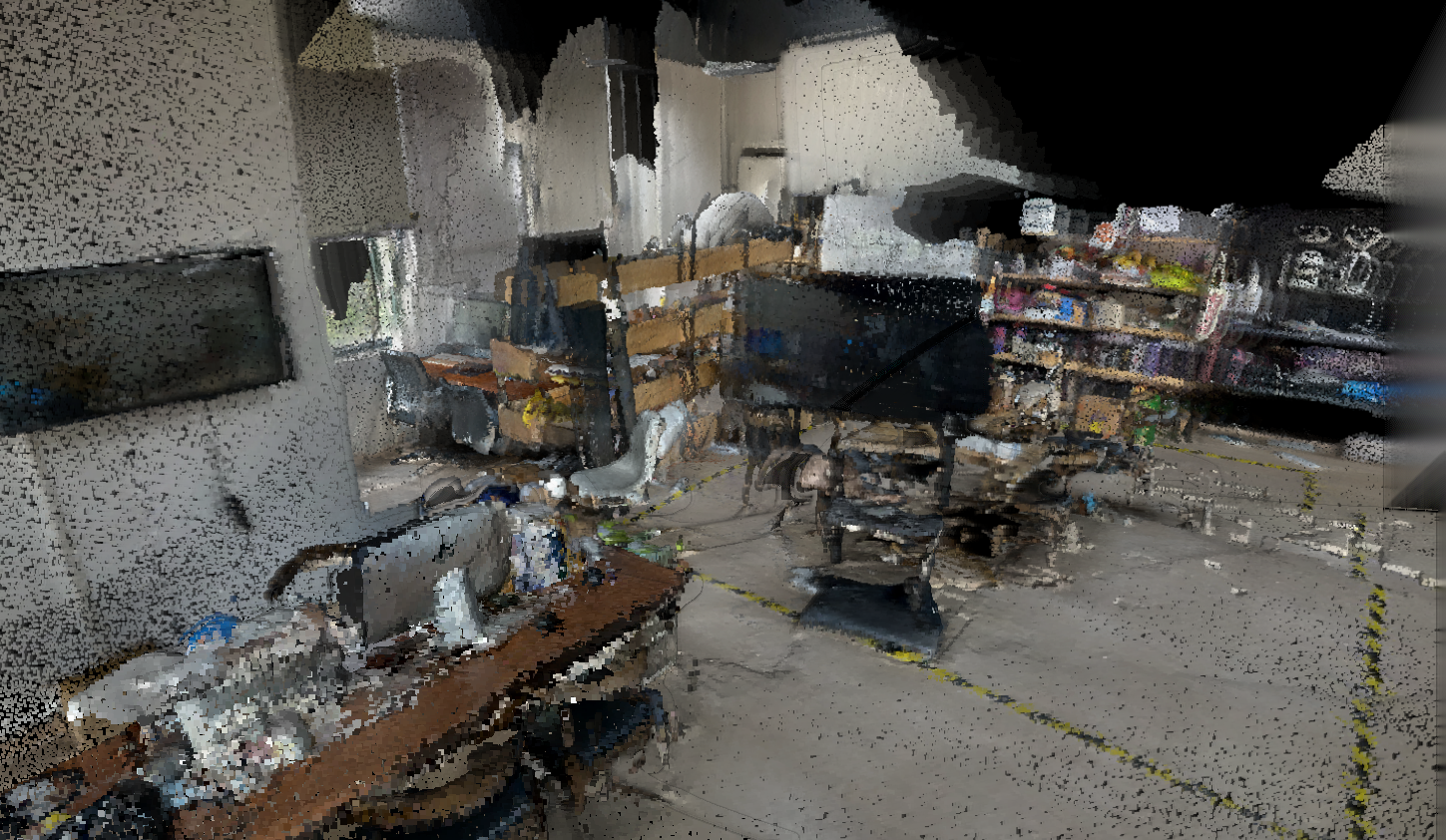} \\
    {\small (c) CPSL Layers (Segmentation)} &
    {\small (d) CPSL Novel View (20° offset)}
\end{tabular}

\caption{\textbf{Full-Scene Volumetric Video as CPSL.}
CPSL decomposes each full-scene frame into depth and semantically aligned layers, enabling parallax-corrected novel-view synthesis while remaining fully 2D-encodable.
}
\label{fig:fsvvd_cpsl_final}
\end{figure}

\subsection{Full-Scene Volumetric Video as CPSL}

To validate that CPSL scales beyond short monocular clips, 
we extend our framework to full-scene volumetric videos using full-scene volumetric video dataset FSVVD~\cite{hu2023fsvvd}. 
Each sequence provides dense multi-camera captures and recorded user trajectories reflecting real exploration behavior in 6-DoF space. We reconstruct per-user view streams by rendering along the recorded trajectories in~\cite{hu2023understanding} and represent them as compact CPSL layer sets. 
For comparison, we also render the same trajectories using point-cloud streaming and adaptive MPI baselines.

As shown in Figure~\ref{fig:fsvvd_cpsl_final}, CPSL preserves
sharp boundaries and stable parallax along long camera
trajectories, while requiring only a fraction of the bandwidth
of conventional point-cloud streaming. 
Table~\ref{tab:fsvvd} reports the quantitative comparison:
CPSL attains comparable perceptual quality (31.2 dB PSNR,
0.083 LPIPS) with over 7× lower bitrate (2.3 Mbps vs.\ 18 Mbps),
and sustains real-time playback.

\section{Conclusion and Discussion}
We presented \textbf{Content-Promoted Scene Layers (CPSL)}, a compact 2.5D video representation that bridges conventional 2D video and volumetric media.
CPSL decomposes monocular video into a small set of depth-ordered, content-aligned RGBA layers, augmented with an edge–depth cache (EDC) that supports the Dynamic Pixel Strip (DPS) to maintain continuous boundaries across views.
By combining geometry-aware layering and temporally coherent propagation, CPSL achieves parallax-consistent novel-view rendering and real-time playback at low bitrate, providing up to 7$\times$ compression over point-cloud streaming while preserving spatial fidelity.

Despite these advantages, CPSL inherits the fronto-parallel assumption of layered representations and may struggle with highly oblique surfaces or fine semi-transparent structures.
Future work will explore arbitrary layer orientations, learning-based adaptive blending, and integration with perceptual streaming protocols for scalable immersive video delivery.

{
    \small
    \bibliographystyle{ieeenat_fullname}
    \bibliography{main}

@String(CVPR= {IEEE Conf. Comput. Vis. Pattern Recog.})

@String(TOG= {ACM Trans. Graph.})

@String(VR   = {Vis. Res.})

@String(CVPR  = {CVPR})

@String(TOG   = {ACM TOG})

@inproceedings{de2025low,
  title={Low-Latency Volumetric Video Conferencing in Congested Networks Through L4S},
  author={De Fr{\'e}, Matthias and van der Hooft, Jeroen and Chang, Chia-Yu and De Schepper, Koen and Alface, Patrice Rondao and De Vleeschauwer, Danny and Wauters, Tim and Steenkiste, Peter and De Turck, Filip},
  booktitle={Proceedings of the 16th ACM Multimedia Systems Conference},
  pages={113--123},
  year={2025}
}

@inproceedings{tucker2020single,
  title={Single-view view synthesis with multiplane images},
  author={Tucker, Richard and Snavely, Noah},
  booktitle={Proceedings of the IEEE/CVF Conference on Computer Vision and Pattern Recognition},
  pages={551--560},
  year={2020}
}

@inproceedings{shade1998layered,
  title={Layered depth images},
  author={Shade, Jonathan and Gortler, Steven and He, Li-wei and Szeliski, Richard},
  booktitle={Proceedings of the 25th annual conference on Computer graphics and interactive techniques},
  pages={231--242},
  year={1998}
}

@inproceedings{hu2023understanding,
  title={Understanding user behavior in volumetric video watching: Dataset, analysis and prediction},
  author={Hu, Kaiyuan and Yang, Haowen and Jin, Yili and Liu, Junhua and Chen, Yongting and Zhang, Miao and Wang, Fangxin},
  booktitle={Proceedings of the 31st ACM International Conference on Multimedia},
  pages={1108--1116},
  year={2023}
}

@article{lombardi2019neural,
  title={Neural volumes: Learning dynamic renderable volumes from images},
  author={Lombardi, Stephen and Simon, Tomas and Saragih, Jason and Schwartz, Gabriel and Lehrmann, Andreas and Sheikh, Yaser},
  journal={arXiv preprint arXiv:1906.07751},
  year={2019}
}

@article{zhou2018stereo,
  title={Stereo magnification: Learning view synthesis using multiplane images},
  author={Zhou, Tinghui and Tucker, Richard and Flynn, John and Fyffe, Graham and Snavely, Noah},
  journal={arXiv preprint arXiv:1805.09817},
  year={2018}
}

@article{hu2025livevv,
  title={LiveVV: Human-Centered Live Volumetric Video Streaming System},
  author={Hu, Kaiyuan and Chen, Yongting and Han, Kaiying and Li, Boyan and Yang, Haowen and Jin, Yili and Liu, Junhua and Wang, Fangxin},
  journal={IEEE Internet of Things Journal},
  year={2025},
  publisher={IEEE}
}

@article{chen2025v2m4,
  title={V2M4: 4D Mesh Animation Reconstruction from a Single Monocular Video},
  author={Chen, Jianqi and Zhang, Biao and Tang, Xiangjun and Wonka, Peter},
  journal={arXiv preprint arXiv:2503.09631},
  year={2025}
}

@inproceedings{han2022single,
  title={Single-view view synthesis in the wild with learned adaptive multiplane images},
  author={Han, Yuxuan and Wang, Ruicheng and Yang, Jiaolong},
  booktitle={ACM SIGGRAPH 2022 Conference Proceedings},
  pages={1--8},
  year={2022}
}

@article{muddala2016spatio,
  title={Spatio-temporal consistent depth-image-based rendering using layered depth image and inpainting},
  author={Muddala, Suryanarayana M and Olsson, Roger and Sj{\"o}str{\"o}m, M{\aa}rten},
  journal={EURASIP Journal on Image and Video Processing},
  volume={2016},
  number={1},
  pages={9},
  year={2016},
  publisher={Springer}
}

@inproceedings{luvizon2021adaptive,
  title={Adaptive multiplane image generation from a single internet picture},
  author={Luvizon, Diogo C and Carvalho, Gustavo Sutter P and dos Santos, Andreza A and Conceicao, Jhonatas S and Flores-Campana, Jose L and Decker, Luis GL and Souza, Marcos R and Pedrini, Helio and Joia, Antonio and Penatti, Otavio AB},
  booktitle={Proceedings of the IEEE/CVF Winter Conference on Applications of Computer Vision},
  pages={2556--2565},
  year={2021}
}

@article{maglo20153d,
  title={3d mesh compression: Survey, comparisons, and emerging trends},
  author={Maglo, Adrien and Lavou{\'e}, Guillaume and Dupont, Florent and Hudelot, C{\'e}line},
  journal={ACM Computing Surveys (CSUR)},
  volume={47},
  number={3},
  pages={1--41},
  year={2015},
  publisher={ACM New York, NY, USA}
}

@inproceedings{wu20244d,
  title={4d gaussian splatting for real-time dynamic scene rendering},
  author={Wu, Guanjun and Yi, Taoran and Fang, Jiemin and Xie, Lingxi and Zhang, Xiaopeng and Wei, Wei and Liu, Wenyu and Tian, Qi and Wang, Xinggang},
  booktitle={Proceedings of the IEEE/CVF conference on computer vision and pattern recognition},
  pages={20310--20320},
  year={2024}
}

@inproceedings{zhang2023structural,
  title={Structural multiplane image: Bridging neural view synthesis and 3d reconstruction},
  author={Zhang, Mingfang and Wang, Jinglu and Li, Xiao and Huang, Yifei and Sato, Yoichi and Lu, Yan},
  booktitle={Proceedings of the IEEE/CVF Conference on Computer Vision and Pattern Recognition},
  pages={16707--16716},
  year={2023}
}

@inproceedings{choi2019extreme,
  title={Extreme view synthesis},
  author={Choi, Inchang and Gallo, Orazio and Troccoli, Alejandro and Kim, Min H and Kautz, Jan},
  booktitle={Proceedings of the IEEE/CVF international conference on computer vision},
  pages={7781--7790},
  year={2019}
}

@inproceedings{teed2020raft,
  title={Raft: Recurrent all-pairs field transforms for optical flow},
  author={Teed, Zachary and Deng, Jia},
  booktitle={European conference on computer vision},
  pages={402--419},
  year={2020},
  organization={Springer}
}

@inproceedings{shih20203d,
  title={3d photography using context-aware layered depth inpainting},
  author={Shih, Meng-Li and Su, Shih-Yang and Kopf, Johannes and Huang, Jia-Bin},
  booktitle={Proceedings of the IEEE/CVF conference on computer vision and pattern recognition},
  pages={8028--8038},
  year={2020}
}

@book{hartley2003multiple,
  title={Multiple view geometry in computer vision},
  author={Hartley, Richard and Zisserman, Andrew},
  year={2003},
  publisher={Cambridge university press}
}

@inproceedings{porter1984compositing,
  title={Compositing digital images},
  author={Porter, Thomas and Duff, Tom},
  booktitle={Proceedings of the 11th annual conference on Computer graphics and interactive techniques},
  pages={253--259},
  year={1984}
}

@inproceedings{peng2023representing,
  title={Representing volumetric videos as dynamic mlp maps},
  author={Peng, Sida and Yan, Yunzhi and Shuai, Qing and Bao, Hujun and Zhou, Xiaowei},
  booktitle={Proceedings of the IEEE/CVF Conference on Computer Vision and Pattern Recognition},
  pages={4252--4262},
  year={2023}
}

@article{jin2023capture,
  title={From capture to display: A survey on volumetric video},
  author={Jin, Yili and Hu, Kaiyuan and Liu, Junhua and Wang, Fangxin and Liu, Xue},
  journal={arXiv preprint arXiv:2309.05658},
  year={2023}
}

@article{wu2024ai,
  title={Ai-enhanced virtual reality in medicine: A comprehensive survey},
  author={Wu, Yixuan and Hu, Kaiyuan and Chen, Danny Z and Wu, Jian},
  journal={arXiv preprint arXiv:2402.03093},
  year={2024}
}

@article{gao2022monocular,
  title={Monocular dynamic view synthesis: A reality check},
  author={Gao, Hang and Li, Ruilong and Tulsiani, Shubham and Russell, Bryan and Kanazawa, Angjoo},
  journal={Advances in Neural Information Processing Systems},
  volume={35},
  pages={33768--33780},
  year={2022}
}

@inproceedings{choi2023tmo,
  title={Tmo: Textured mesh acquisition of objects with a mobile device by using differentiable rendering},
  author={Choi, Jaehoon and Jung, Dongki and Lee, Taejae and Kim, Sangwook and Jung, Youngdong and Manocha, Dinesh and Lee, Donghwan},
  booktitle={Proceedings of the IEEE/CVF Conference on Computer Vision and Pattern Recognition},
  pages={16674--16684},
  year={2023}
}

@article{sze2014high,
  title={High efficiency video coding (HEVC)},
  author={Sze, Vivienne and Budagavi, Madhukar and Sullivan, Gary J},
  journal={Integrated circuit and systems, algorithms and architectures},
  volume={39},
  pages={40},
  year={2014},
  publisher={Springer}
}

@inproceedings{li2022neural,
  title={Neural 3d video synthesis from multi-view video},
  author={Li, Tianye and Slavcheva, Mira and Zollhoefer, Michael and Green, Simon and Lassner, Christoph and Kim, Changil and Schmidt, Tanner and Lovegrove, Steven and Goesele, Michael and Newcombe, Richard and others},
  booktitle={Proceedings of the IEEE/CVF conference on computer vision and pattern recognition},
  pages={5521--5531},
  year={2022}
}

@inproceedings{pumarola2021d,
  title={D-nerf: Neural radiance fields for dynamic scenes},
  author={Pumarola, Albert and Corona, Enric and Pons-Moll, Gerard and Moreno-Noguer, Francesc},
  booktitle={Proceedings of the IEEE/CVF conference on computer vision and pattern recognition},
  pages={10318--10327},
  year={2021}
}

@article{artois2024opendibr,
  title={OpenDIBR: open real-time depth-image-based renderer of light field videos for VR},
  author={Artois, Julie and Courteaux, Martijn and Van Wallendael, Glenn and Lambert, Peter},
  journal={Multimedia Tools and Applications},
  volume={83},
  number={9},
  pages={25797--25815},
  year={2024},
  publisher={Springer}
}

@article{wiegand2003overview,
  title={Overview of the H. 264/AVC video coding standard},
  author={Wiegand, Thomas and Sullivan, Gary J and Bjontegaard, Gisle and Luthra, Ajay},
  journal={IEEE Transactions on circuits and systems for video technology},
  volume={13},
  number={7},
  pages={560--576},
  year={2003},
  publisher={IEEE}
}

@article{yariv2021volume,
  title={Volume rendering of neural implicit surfaces},
  author={Yariv, Lior and Gu, Jiatao and Kasten, Yoni and Lipman, Yaron},
  journal={Advances in neural information processing systems},
  volume={34},
  pages={4805--4815},
  year={2021}
}

@article{muller2022instant,
  title={Instant neural graphics primitives with a multiresolution hash encoding},
  author={M{\"u}ller, Thomas and Evans, Alex and Schied, Christoph and Keller, Alexander},
  journal={ACM transactions on graphics (TOG)},
  volume={41},
  number={4},
  pages={1--15},
  year={2022},
  publisher={ACM New York, NY, USA}
}

@article{mildenhall2020nerf,
  title = {NeRF: Representing Scenes as Neural Radiance Fields for View Synthesis},
  author = {Mildenhall, Ben and Srinivasan, Pratul P. and Tancik, Matthew and Barron, Jonathan T. and Ramamoorthi, Ravi and Ng, Ren},
  journal = {arXiv preprint arXiv:2003.08934},
  year = {2020},
  url = {https://arxiv.org/abs/2003.08934}
}

@inproceedings{pumarola2021dnerf,
  title = {D-NeRF: Neural Radiance Fields for Dynamic Scenes},
  author = {Pumarola, Albert and Agudo, Alberto and Serr{\`a}, Jan and Sanfeliu, Alberto and Moreno-Noguer, Francesc},
  booktitle = {Proceedings of the IEEE/CVF Conference on Computer Vision and Pattern Recognition (CVPR)},
  year = {2021},
  pages = {10365--10374},
  url = {https://openaccess.thecvf.com/content/CVPR2021/html/Pumarola_D-NeRF_Neural_Radiance_Fields_for_Dynamic_Scenes_CVPR_2021_paper.html}
}

@inproceedings{zhong2024low,
  title={Low-bitrate Volumetric Video Streaming with Depth Image},
  author={Zhong, Junquan and Zhang, Haodan and Jia, Quanlu and Wu, Jiangkai and Wang, Peiheng and Wang, Haoyang and Liu, Liming and Zhang, Xinggong and Guo, Zongming},
  booktitle={Proceedings of the 2024 SIGCOMM Workshop on Emerging Multimedia Systems},
  pages={39--44},
  year={2024}
}

@inproceedings{hu2023fsvvd,
  title={FSVVD: A dataset of full scene volumetric video},
  author={Hu, Kaiyuan and Jin, Yili and Yang, Haowen and Liu, Junhua and Wang, Fangxin},
  booktitle={Proceedings of the 14th Conference on ACM Multimedia Systems},
  pages={410--415},
  year={2023}
}

@inproceedings{wang2025shape,
  title={Shape of motion: 4d reconstruction from a single video},
  author={Wang, Qianqian and Ye, Vickie and Gao, Hang and Zeng, Weijia and Austin, Jake and Li, Zhengqi and Kanazawa, Angjoo},
  booktitle={Proceedings of the IEEE/CVF International Conference on Computer Vision},
  pages={9660--9672},
  year={2025}
}

@inproceedings{kerbl2023gaussiansplatting,
  title = {3D Gaussian Splatting for Real-Time Radiance Field Rendering},
  author = {Kerbl, Bernhard and Kopanas, Georgios and Leimk{\"u}hler, Thomas and Drettakis, George},
  booktitle = {Proceedings of ACM SIGGRAPH / ACM Transactions on Graphics},
  year = {2023},
  url = {https://arxiv.org/abs/2308.04079},
  note = {Published as SIGGRAPH 2023}
}

@inproceedings{shi2023vpcc,
  title = {Enabling Low Bit-Rate MPEG V-PCC-encoded Volumetric Video Streaming with 3D Sub-sampling},
  author = {Shi, Yuang and Venkatram, Pranav and Ding, Yifan and Ooi, Wei Tsang},
  booktitle = {Proceedings of the ACM on Multimedia Systems (MMSys)},
  year = {2023},
  pages = {108--118},
  doi = {10.1145/3587819.3590981},
  url = {https://doi.org/10.1145/3587819.3590981}
}
}


\end{document}